\documentclass[10pt, a4paper]{article}
\usepackage{lrec}
\usepackage{graphicx}
\usepackage{tabularx}
\usepackage{soul}

\usepackage{epstopdf}
\usepackage[latin1]{inputenc}

\usepackage{hyperref}
\usepackage{xstring}

\usepackage{amssymb}
\usepackage{graphicx}
\usepackage{times}
\usepackage{latexsym}
\usepackage{todonotes}
\usepackage{dcolumn}
\usepackage{graphics}
\usepackage{url}
\usepackage{longtable}
\usepackage{supertabular}
\usepackage{pdflscape}
\usepackage{pifont}

\usepackage{rotating}
\usepackage{tabularx}
\usepackage{lscape}
\usepackage{multirow}
\usepackage{multicol}
\usepackage{float}
\usepackage{footnote}
\usepackage{verbatim}
\usepackage{hyperref}

\usepackage{url}
\usepackage{mathtools}
\usepackage{scrextend}
\usepackage{xspace}
\usepackage{listingsutf8}
\usepackage{color}
\usepackage{booktabs}
\usepackage{tikz}
\usepackage{lipsum}
\usepackage[ruled,vlined]{algorithm2e}
\usepackage{mdframed}
\usepackage{wrapfig}
\usepackage[nolist]{acronym}
\usepackage{adjustbox}
\usepackage{array}
\usepackage{subfigure}
\usepackage{caption}

\definecolor{gray}{rgb}{0.4,0.4,0.4}
\definecolor{darkblue}{rgb}{0.0,0.0,0.6}
\definecolor{cyan}{rgb}{0.0,0.6,0.6}
\definecolor{cffffff}{RGB}{255,255,255}

\lstdefinelanguage{XML}
{
  morestring=[b]",
  morestring=[s]{>}{<},
  morecomment=[s]{<?}{?>},
  stringstyle=\color{black},
  identifierstyle=\color{darkblue},
  keywordstyle=\color{cyan},
  morekeywords={xmlns,version,type}
}

\definecolor{codegreen}{rgb}{0,0.6,0}
\definecolor{codegray}{rgb}{0.5,0.5,0.5}
\definecolor{codeblue}{rgb}{0,0,255}
\definecolor{backcolour}{rgb}{0.95,0.95,0.92}

\lstdefinestyle{rdf}{numberblanklines=true, morekeywords={},
backgroundcolor=\color{backcolour},   
    commentstyle=\color{codegreen},
    keywordstyle=\color{magenta},
    numberstyle=\tiny\color{codegray},
    stringstyle=\color{codeblue},
    basicstyle=\footnotesize,
    breakatwhitespace=false,         
    breaklines=true,                 
    captionpos=b,                    
    keepspaces=true,                 
    numbers=left,                    
    numbersep=5pt,                  
    showspaces=false,                
    showstringspaces=false,
    showtabs=false,                  
    tabsize=2
}

\newcolumntype{R}[2]{%
    >{\adjustbox{angle=#1,lap=\width-(#2)}\bgroup}%
    l%
    <{\egroup}%
}

\lstdefinestyle{sparql}{numberblanklines=true, morekeywords={SERVICE,SELECT,DISTINCT,SAMPLE,FROM,WHERE,FILTER,ORDER,GROUP,BY,IN,AS,LIMIT}}
 
\newcolumntype{d}[1]{D{.}{.}{#1}}
\newcommand{\tablefont}[1]{\fontsize{3mm}{3.2mm}\selectfont}

\newcommand{\executeiffilenewer}[3]{%
 \ifnum\pdfstrcmp{\pdffilemoddate{#1}}%
 {\pdffilemoddate{#2}}>0%
 {\immediate\write18{#3}}\fi%
}

\newcommand{\lids}{\textsc{LIdioms}\xspace}
\newcommand{\limes}{\textsc{Limes}\xspace}

\makeatletter
\newcommand\footnoteref[1]{\protected@xdef\@thefnmark{\ref{#1}}\@footnotemark}
\makeatother

\title{\lids: A Multilingual Linked Idioms Data Set}

\name{Diego Moussallem\textsuperscript{1,2}, Mohamed Ahmed Sherif\textsuperscript{2},Diego Esteves\textsuperscript{1} Marcos Zampieri\textsuperscript{3},\\
\large{\textbf{Axel-Cyrille Ngonga Ngomo\textsuperscript{2}}}}

\address{%
    \textsuperscript{1}Faculty of Mathematics and Computer Science - University of Leipzig, Germany\\
    \textsuperscript{2}Data Science Group - University of Paderborn, Germany\\
    \textsuperscript{3}Research Group in Computational Linguistics - University of Wolverhampton, United Kingdom \\
    lastname@informatik.uni-leipzig.de\\
}

\begin{acronym}[UML]
	\acro{AOS}{Agricultural Ontology Services}
	\acro{AGRIS}{Agricultural Science and Technology}
	\acro{API}{Application Programming Interface}
	\acro{BPSO}{Binary Particle-Swarm Optimization}
	\acro{BPMLOD}{Best Practices for Multilingual Linked Open Data}
	\acro{CBD}{Concise Bounded Description}
	\acro{COG}{Content Oriented Guidelines}
	\acro{CSV}{Comma-Separated Values}
	\acro{DPSO}{Deterministic Particle-Swarm Optimization}
	\acro{DALY}{Disability Adjusted Life Year}

	\acro{ER}{Entity Resolution}
	\acro{EM}{Expectation Maximization}
	\acro{FAO}{Food and Agriculture Organization of the United Nations}
	\acro{GIS}{Geographic Information Systems}
	\acro{GHO}{Global Health Observatory}
	\acro{HDI}{Human Development Index}
	\acro{ICT}{Information and communication technologies}
	\acro{LR}  {Language Resource}
	\acro{LD}  {Linked Data}
	\acro{LLOD}  {Linguistic Linked Open Data}
	\acro{LIMES}{LInk discovery framework for MEtric Spaces}
	\acro{LS}  {Link Specifications}
	\acro{LDIF}{Linked Data Integration Framework}
	\acro{LGD} {LinkedGeoData}
	\acro{LOD} {Linked Open Data}
	\acro{MSE}{Mean Squared Error}
	\acro{MWE}{Multiword Expressions}
	\acro{MT}{Machine Translation}
	\acro{NIF}{Natural Language Processing Interchange Format}
	\acro{NIF4OGGD}{NLP Interchange Format for Open German Governmental Data}
	\acro{NLP}{Natural Language Processing}
	\acro{OSM}{OpenStreetMap}
	\acro{OWL}{Web Ontology Language}
	\acro{PFM}{Pseudo-F-Measures}
	\acro{PSO}{Particle-Swarm Optimization}
	\acro{QA}{Question Answering}
	\acro{RDF}{Resource Description Framework}
	\acro{SKOS}{Simple Knowledge Organization System}
	\acro{SPARQL}{SPARQL Protocol and RDF Query Language}
	\acro{SRL}{Statistical Relational Learning}
	\acro{SWT}{Semantic Web Technologies}
	\acro{UML}{Unified Modeling Language}
	\acro{WHO}{World Health Organization}
	\acro{WKT}{Well-Known Text}
	\acro{W3C}{World Wide Web Consortium}
	\acro{YPLL}{Years of Potential Life Lost}
\end{acronym}  

\abstract{
In this paper, we describe the \lids data set, a multilingual RDF representation of idioms currently containing five languages: English, German, Italian, Portuguese, and Russian. 
The data set is intended to support natural language processing applications by providing links between idioms across languages. The underlying data was crawled and integrated from various sources. To ensure the quality of the crawled data, all idioms were evaluated by at least two native speakers. Herein, we present the model devised for structuring the data. We also provide the details of linking \lids to well-known multilingual data sets such as BabelNet. The resulting data set complies with best practices according to Linguistic Linked Open Data Community.\\ \newline \Keywords{multilingual, idioms, translation} }

\begin{document}

\maketitleabstract

\section{Introduction}
\label{sec:intro}
Recently, the \ac{LLOD}\footnote{\url{http://linguistics.okfn.org/}} movement has gained significant momentum.
According to~\newcite{mccrae2016open}, a large number of linguistic data sets have been extracted from various sources and been represented as \ac{LD}. This new movement was motivated by the novel capabilities of the \ac{LD} paradigm pertaining to transforming, sharing, and linking linguistic data on the Web~\cite{chiarcos2012linked}. Resources such as dictionaries and knowledge bases are essential in the development of \ac{NLP} systems. However, most of these resources are still bilingual on the \ac{LLOD}. Thus, becoming worthwhile to develop multilingual knowledge bases by reusing these bilingual contents. Multilingualism is important not only for sharing information across Web but also for learning new concepts from other cultures. 

There are many data sets and linguistic resources available at \ac{LLOD}, however, most of them do not contain much information about \ac{MWE}. \ac{MWE} are known to constitute a difficult problem on a number of \ac{NLP} tasks such as machine translation, language generation, and sentiment analysis/opinion mining. There are different types of \ac{MWE}, according to~\newcite{nunberg1994idioms}, \ac{MWE} are categorized as phrase verbs, compounds, fixed expression, semi-fixed expressions, idioms, slang, and others. This work focuses on idioms, a particular type of \ac{MWE}.

Most idioms are culture-bound and their senses come from particular concepts of everyday life to a given culture. By definition, idioms are a sequence of words whose meaning cannot be derived from the meaning of words that constitute them~\cite{nunberg1994idioms}. 
Idioms are generally classified as non-com\-po\-si\-ti\-o\-nal. One of the direct consequences of non-compositionality is the impossibility of translating this kind of word group literally~\cite{nunberg1994idioms} posing challenges to human translators and to machine translation systems. 

In this paper, we propose \lids, a multilingual linked data set of idioms in five languages. In \lids, we do not distinguish between idioms sub-categories and thus work on idioms in general by providing lexical and semantic knowledge on a multilingual basis. The selected languages are English, German, Italian, Portuguese, and Russian. This choice of languages intends to show the possibility of correct translations among idioms independent of their language family, syntax or culture. Additionally, one of the goals of \lids is to support further investigations of similarity among idioms from different languages.

In the following, we begin by presenting the related work (\autoref{sec:relatedwork}) and the data sources that we used for the extraction (\autoref{sec:source}). In \autoref{sec:onto}, we give an overview of the model that underlies our data set. \autoref{sec:generation} depicts the creation process that led to the publication of our data set. 
In \autoref{sec:linking}, we present our approach to link \lids internally and externally. Then, we present usage scenarios for our data set in \autoref{sec:uses}. Subsequently, we discuss \ac{LLOD} quality in \autoref{sec:discussion} and we conclude the paper and provide avenues for future work in \autoref{sec:conclusion}.


\section{Related Work}
\label{sec:relatedwork}

A large number of ontologies have been developed to represent natural language data as \ac{LD} on the Web of Data. In this context, the well-known ontology {\it{lemon}}~\cite{mccrae2012interchanging} was originally developed to model lexical data in mono or multilingual way. Subsequently, a significant amount of effort has been invested in order to improve the support of multilingual contents. To this end, other modules have been extended from {\it{lemon}} for representing multilingual data including~\cite{gracia2014enabling}, which extends some of the {\it{lemon}} properties describing relationships among translations.

Recently, multilingual data sets have been created such as DBnary~\cite{serasset2012dbnary}, which was released with the main purpose of describing translations among lexical entries. Another resource that describes multilingual content is BabelNet~\cite{navigli2010babelnet}, which integrates knowledge from various lexical resources, such as WordNet~\cite{miller1995wordnet}. Additionally, BabelNet has adopted the {\it{lemon}} structure for representing lexical entries~\cite{ehrmann2014representing}. Although these resources are linked lexical multilingual data sets, they contain a limited number of idioms described correctly along with their respective translations across languages. This lack of information about \ac{MWE} and idioms is due to the missing appropriate ontologies and vocabularies for handling this phenomena properly. Despite \emph{Lexinfo} ontology~\cite{cimiano2011lexinfo} contains a certain property just for representing idioms, there are no appropriate classes to reuse this information. Fortunately, the W3C Ontology Lexica Community Group\footnote{\url{https://www.w3.org/community/ontolex/}} has created an extension of {\it{lemon}} called \emph{Ontolex}\footnote{\url{https://www.w3.org/community/ontolex/wiki/Final_Model_Specification}} in order to not only address this lack of information but also to describe more appropriately linguistic terms~\cite{bosque2015applying}. 
Thus, enabling \lids to represent a particular type of linguistic unit, that is to say idioms. In the following, we present the data set creation process in more detail.

Additionally, a number of multilingual data sets have been published as \ac{LOD} in the last years. The well-known knowledge base of \emph{DBpedia}~\cite{lehmann2015dbpedia} is one of first multilingual knowledge bases extracted from \emph{Wikipedia}\footnote{\url{https://www.wikipedia.org/}}. Recently, the \emph{Semantic Quran} data set has published translations of the Quran in $43$ different languages as linked data~\cite{sherif2015semantic}. \emph{xLiD-Lexica}~\cite{zhang2014xlid} is a cross-lingual linked data lexica which is constructed by exploiting all language versions of Wikipedia. \emph{Terminesp}~\cite{bosque2015applying} is another multilingual resource for terms along with their definitions in various languages.


\section{Data Sources}
\label{sec:source}
In this section, we list the data sources from which \lids originates, where we describe the data collection process of each data source. In addition, we discuss how we ensure the quality of the collected data.

\subsection{Data sets}

We collected a set of \ac{MWE} from the online lexical resources: (1) \emph{Phrase finder}, (2) \emph{Memrise}, (3) \emph{Collins} and (4) \emph{Oxford} dictionaries\footnote{\label{all}All repositories web pages \url{http://faturl.com/repositories/?open}}.
\emph{Phrase finder} is an online dictionary about idiomatic expressions created by \emph{Gary Martins}~\cite{martin2007phrase} in 1997 for supporting his post-graduate research in computational linguistics.
\emph{Memrise} is an online course about idiomatic expressions for achieving a native speaker level.
 \emph{Collins} and \emph{Oxford} provide high quality lexical resources.
 Therefore, we use them to guarantee the quality of the idioms definitions and also for gathering some additional idioms. \emph{Memrise} and \emph{Oxford} provided idioms in English, German, Italian, and Russian languages, while the idioms in \emph{Phrase finder} and \emph{Collins} are in English. 
 The Portuguese idioms were initially gathered from \emph{Wikipedia} Portuguese page\footnoteref{all} but because of the limited number of the available Portuguese idioms in Wikipedia, we asked four native speakers (one from Portugal and the other three from Brazil) to add more Portuguese idioms.
 
For the sake of clarity pertaining to the copyrights to use the data, \emph{Memrise} and \emph{Collins} granted us a full permission while the others data providers have a free licence policy when to use the data for research purposes.

\subsection{Data Collection process}
\label{sec:dataCollectionProcess}

Using a custom web crawler, we collected the \ac{MWE} from these aforementioned on-line data sources.
Each of the crawled resources has specific pages about each \ac{MWE}, which ease the configuration of our crawler.
Note that, all data sources are bilingual but not necessarily including English as one of the involved languages.
For instance, \emph{Oxford} has idiomatic expressions from Italian to Portuguese. We also noticed that most on-line dictionaries does not correctly categorize \ac{MWE}. For example, in some cases the meaning of \ac{MWE} can
be deduced from the meaning of their components (e.g. \emph{``by the book"}) while in other cases this is not possible (e.g. \emph{``out of the blue"}). Therefore, \ac{MWE} which can be represented by the meaning of their components should not be into the same category as the others with pragmatic meanings (i.e. non-compositional idioms). 

Collecting the right idioms was a hard task due to the lack of \ac{MWE} categorization.
Thus, we carried out the idiom collection manually where we discarded all the entries that were semantically equivalent to their lexical definitions which means to be \emph{not} non-compositional.
We dubbed this process \emph{pragmatically-based selection}.
The pragmatically-based selection identified only $50\%$ of the \ac{MWE} retrieved by our crawler as idioms. For instance, the idiom mentioned before \emph{``by the book"} means ``to follow the rules as demand". The meaning of ``book" is ``a stuff which contains information, rules, descriptions, and it can be a manual". Therefore, this \ac{MWE} is deductible from the meaning of each of its components, the meaning gets ``to follow the book's writing". Therefore, it is not considered an idiom, in contrast of the idiom \emph{``out of the blue"} which means ``an event that occurs unexpectedly", the meaning of ``blue" is ``color" then no relationship exists between ``blue" or ``out of" with ``unexpected happening". 

Moreover, considering that the meaning of idioms may vary according to the geographical location where they are used~\cite{martin2007phrase}. For example, American idioms which come from The United States of America differ from the British idioms which come from United Kingdom. We consider the location of idioms as an important characteristic to be included in \lids.

\subsection{Data Evaluation}

To ensure the quality of the retrieved data, we asked two native speakers and one linguist (per language) to evaluate the extracted idioms and their respective definitions in English.
For evaluating an idiom, each native speaker separately evaluated the idioms' definition. 
Idioms with accepted definitions by both evaluators are accepted. Also, idioms with idioms' definitions marked as wrong by both evaluators were discarded.
In case a mismatch evaluation happens, the idiom was judged by the linguist. 
This procedure resulted in a manually checked data set containing a large number of idioms as shown in Table~\ref{tbl:gold}. 
The \emph{Collection} column shows the number of all \ac{MWE} retrieved by our web crawler.
The \emph{Filter} column shows the number of idioms retrieved based on our pragmatically-based selection, a step which recognizes only idioms among \ac{MWE} (see \autoref{sec:dataCollectionProcess}).
The \emph{Total} column presents the resulting number of idioms after the manual review process made by the natives and the linguist.

\begin{table}[htb]
\begin{tabularx}{0.48\textwidth}{@{}p{0.13\textwidth}p{0.128\textwidth}p{0.1\textwidth}|p{0.5\textwidth}@{}}
 \toprule
\textbf{Language} & \textbf{Collection} & \textbf{Filter} & \textbf{Total}\\
\midrule
  English & 1230 & 600 & 291 \\
  Portuguese & 600 & 215 & 114 \\
  Italian & 500 & 284 & 175\\
  German & 400 & 245 & 130\\
  Russian & 220 & 150 & 105\\
 \bottomrule
\end{tabularx}
\caption{Number of idioms retrieved by step}
\label{tbl:gold}
\end{table}
\section{Semantic Representation Model}
\label{sec:onto}

The representation model of \lids aims at describing idioms correctly as a sub-type of \ac{MWE} together with their translations and geographical usage area. 
For this purpose, \lids data set is based on Ontolex model. We chose the \emph{Ontolex} model because it contains the necessary classes to represent \ac{MWE} and its translations properly. \emph{Ontolex} also reuses the well-known \emph{Lexinfo} ontology which has an essential term type called \texttt{lexinfo:idiom} for representing idioms as one type of \ac{MWE}. 

We used the core Ontolex's classes to model \lids, where (1) we use the class \texttt{on\-to\-lex:Le\-xi\-cal\-En\-try} for representing a lexical entry (i.e. a word, a multi-word expression or an affix), (2) the sub-class \texttt{ontolex:Multi\-word\-Ex\-pres\-sion} is used to specify a lexical entry as a multi-word expression, (3) the  \texttt{on\-to\-lex:Le\-xi\-cal\-Con\-cept} class suits perfectly for representing idioms meaning as its formal definition comprises of ``to be a mental abstraction, concept or a thought that can be described by a given collection of senses". (4) the \texttt{on\-to\-lex:Le\-xi\-cal\-Sen\-se} class for lexical sense of an idiom. (5) the \texttt{on\-to\-lex:Form} class describes the written and alternative forms of the entries and (6) \texttt{on\-to\-lex:Lexicon} class is used for representing a collection of lexical entries.

For translations, \emph{Ontolex} uses the \emph{vartrans} module which connects \texttt{on\-to\-lex:Le\-xi\-cal\-Sense} instances among themselves through \texttt{var\-trans:\-Trans\-la\-tion} class. The \texttt{var\-trans:Trans\-la\-tion} uses the property \texttt{var\-trans:ca\-te\-gory} for describing translations and also representing variations of these translations across entries in the same\footnote{Same entry from a given language with different meanings} or different languages. 
The \texttt{var\-trans} module was inspired by~\cite{gracia2014enabling} and we also reuse one of its translation categories called \texttt{tr\-cat:cul\-tu\-ral\-Equi\-va\-lent} which represents a translation between two entries that are not semantically but pragmatically equivalent. Note that a cultural translation of an idiom is not a literal translation, rather it represents the specific cultural semantics of that idiom. 

For the geographical area of idioms, we use the \texttt{lexvo:usedIn} class from the \emph{Lexvo} Ontology~\cite{de2015lexvo}. The geographical area of an idiom is of great importance because the meaning of an idiom can vary in the same language depending on where it is used (diatopic variation). For instance, the Portuguese idiom \emph{``amarrar o burro"}(its literal translation: ``tie the donkey") means ``to relax" in Portugal while in Brazil it means ``to advise someone about future problems from one action". Furthermore, this idiom has also more meanings even within Brazil, for example, ``to be angry when someone does not allow you to do something" that is typical for children. In addition, some idioms are not understood in all countries even sharing the same language. For instance, the Portuguese idiom \emph{``comprei um mamao"} (eng: ``buy a lemon") is used in Brazil but not in Portugal.

\begin{figure*}[tb]
 \centering
 \includegraphics[width=\textwidth]{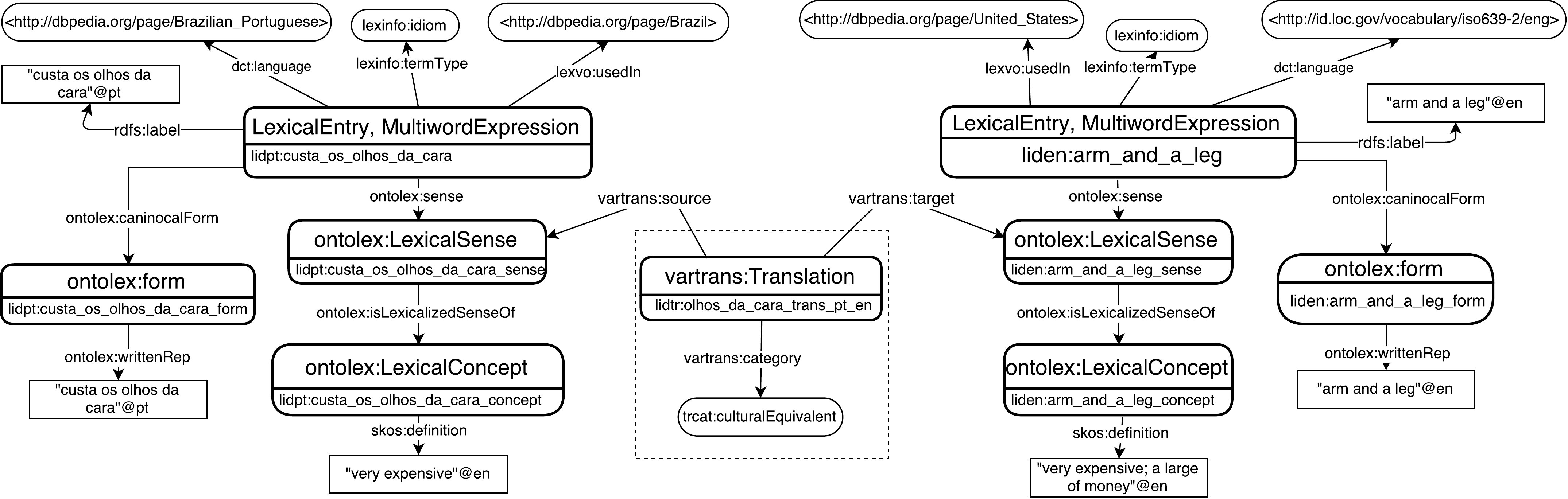}
 \caption{RDF representation of translation of two idioms from Portuguese (``Custa os olhos da cara") to English (``arm and a leg") by entries modeled with the \lids model.}
 \label{fig:modeling}
 \end{figure*}

 In \autoref{fig:modeling}, we present a complete example of a translation of two idioms from Portuguese (``custa os olhos da cara") to English (``arm and a leg") using \emph{vartrans} class  along with the others descriptions modeled by \emph{Ontolex} in \lids.

In order to represent the names of the languages in a unified way, we publish \lids based on the best practices of the International Organization of Standardization (ISO). Given the fact that Brazilian Portuguese does not have an ISO resource, we chose to use the Brazilian Portuguese DBpedia resource\footnote{\url{http://dbpedia.org/page/Brazilian_Portuguese}} for substituting that missing ISO.


\section{RDF Generation}
\label{sec:generation}
The original idioms were crawled in heterogeneous formats such as CSV, XML, and HTML. 
To convert the idiom data into RDF, we used \emph{OpenRefine}\footnote{\url{http://openrefine.org/}} together with its RDF extension. The model underlying the RDF conversion relies on the group of patterns to generate linguistic resources as LD recommended by the \ac{BPMLOD} W3C community group\footnote{\url{https://www.w3.org/community/bpmlod/}}. In spite of our work being multilingual, we followed the patterns for bilingual dictionaries\footnote{\url{https://www.w3.org/2015/09/bpmlod-reports/bilingual-dictionaries/}}. We were able to use bilingual patterns because we use English as pivot language given that all the target translations are in English. 
Thus, the multilingual translations were found by inference relying on the reflexivity property of the \texttt{vartrans:target}. 
For more details about \lids see Table~\ref{tbl:tech} and visit \lids GitLab repository\footnote{\url{https://github.com/dice-group/LIdioms}}.

\begin{table}[tb]
 \begin{tabularx}{0.48\textwidth}{@{}p{0.1\textwidth}p{0.38\textwidth}@{}}
 \toprule
 \textbf{Name}		  & \lids\\
 \textbf{Example}		  & \url{http://lid.aksw.org/en/kill_two_birds_with_one_stone} \\
 \textbf{Dump}		  & \url{http://lid.tabsolucoes.com/dataset/dump-1-0.tar.gz} \\
 \textbf{Sparql}		  & \url{http://lid.aksw.org/sparql} \\
 \textbf{Repository}		  & \url{https://datahub.io/dataset/lidioms} \\
 \textbf{Ver. Date}		  & 20.04.2017 \\
 \textbf{Ver.No}		  & 1.0 \\
 \textbf{License}		  & CC BY-NC-SA 3.0 \\
 \bottomrule
 \end{tabularx}
  \caption{Technical Details \lids.}
    \label{tbl:tech}
 \end{table}

\section{Linking}
\label{sec:linking}
In this section, we describe how we link idioms in \lids internally (i.e. within the data set) and externally (i.e. with other data sets).

\subsection{Internal linking}

\begin{figure}[htb]
\centering
\includegraphics[scale=0.4]{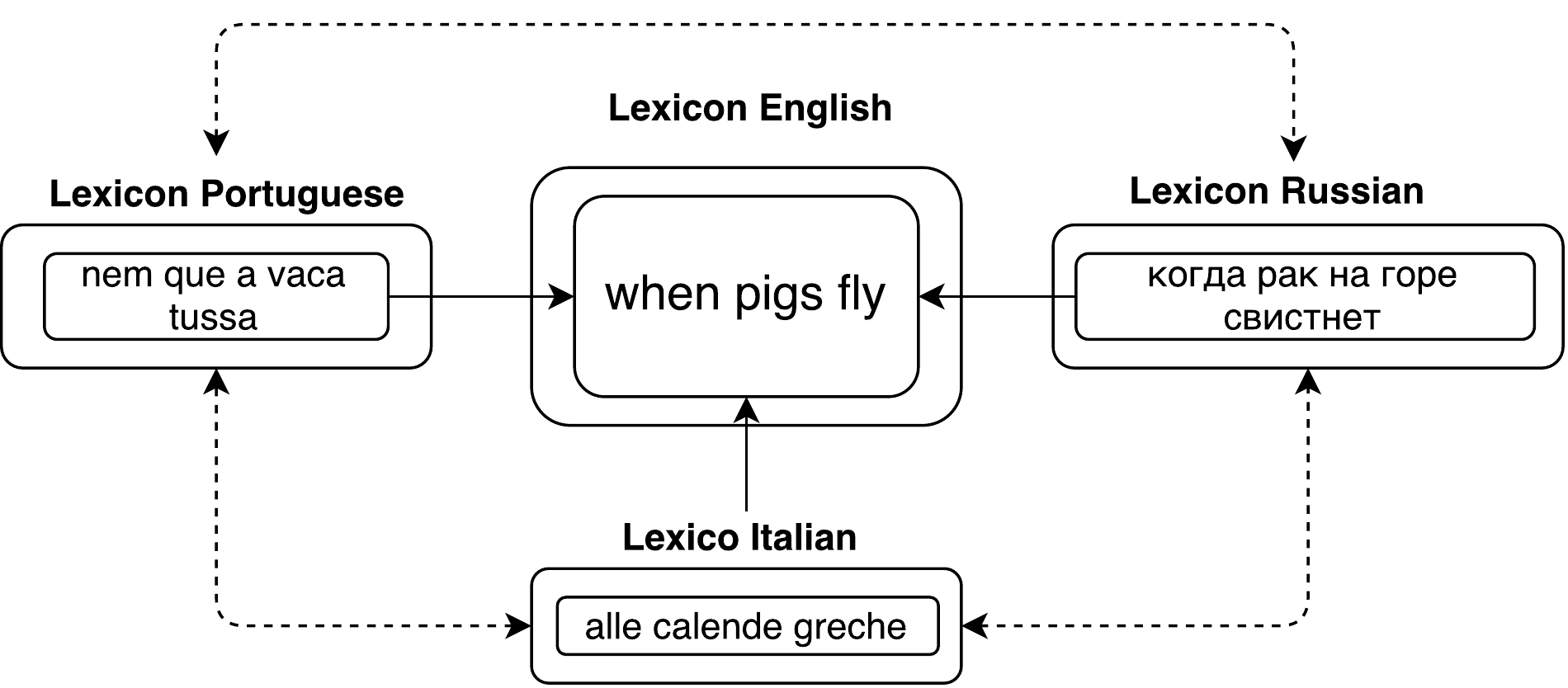}
\caption{An indirect translation excerpt}
\label{fig:equivalent}
\end{figure}

While most of the definitions of the retrieved idioms were in English ($87\%$), only in a few cases the definition was in another language.
We then decided to provide the definitions of all idioms in English regardless of the idioms' original language. 
The other $13\%$ of idioms which had the definitions in another language were translated by a native speaker to English. Therefore, the English definitions became our pivot language, i.e. the idioms' English definitions were used as bridge for the internal linking process across languages. For instance, the \emph{``when pigs fly"} English idiom has the definition ``something that will never happen". In Portuguese, the idiom \emph{``nem que a vaca tussa"} has exactly the same lexical definition, but its literal translation would be ``nor the cow cough". 
Still, it is valid to decide to link these two idioms internally based on their definitions.
\autoref{fig:equivalent} illustrates the main idea underlying this work, i.e. the provision of indirect translations (represented by dotted line) of idioms through a pivot language.

Note that, some idioms have multiple idiomatic equivalents in other languages while others have none.
However, some idioms have definitions with almost equivalent syntactic structures while the semantics of the definitions are very different. For instance, the English idiom \emph{``Once in a blue moon"} means \emph{``something that happens rarely"} and another English idiom \emph{``When pigs fly"} means \emph{``something that will never happen"}. This kind of phenomena is likely to decrease the quality of an automatic linking process, because current link discovery frameworks~\cite{nentwig2015survey} only support syntax-based string similarities. Given the lack of support of semantic-based string similarity functions, the internal linking was carried out \emph{manually} by the authors and a cross-validation among the natives and linguists were done on this manual internal linking.

The \autoref{tbl:trans} shows the number of direct and indirect translations found for the selected idioms per language.

\begin{table}[!htb]
\begin{tabularx}{0.48\textwidth}{@{}p{0.1\textwidth}p{0.03\textwidth}p{0.03\textwidth}p{0.03\textwidth}p{0.03\textwidth}p{0.03\textwidth}|p{0.03\textwidth}@{}}
 \toprule
& \textbf{EN} & \textbf{PT} & \textbf{IT} & \textbf{DE} & \textbf{RU} & \textbf{Total} \\
\midrule
  \textbf{Idioms} & 291 & 114 & 175 & 130 & 105 & 815 \\
  \midrule
  \textbf{Translations} & 192 & 79 & 73 & 62 & 82 & 488 \\
 \bottomrule
\end{tabularx}
\caption{Number of idioms and Translations}
\label{tbl:trans}
\end{table}

\subsection{External linking}

\begin{table*}[!htb]
\centering
\small
\resizebox{\textwidth}{!}{
\begin{tabularx}{\textwidth}{@{} XXXXXXXX @{}} 
\toprule
\textbf{Languages} & \textbf{\lids} & \textbf{BabelNet Retrieval} & \textbf{BabelNet Accepted} & \textbf{BabelNet Precision} & \textbf{DBnary Retrieval} & \textbf{DBnary Accepted} & \textbf{DBnary Precision}\\
\midrule
 English & 291 & 600 & 195 & 0.325 & 362 & 323 & 0.892  \\
 Portuguese & 114 & 23 & 9 & 0.391 & 26 & 4 & 0.153 \\
 Italian & 175 & 52 & 33 & 0.634 & 4 & 4 & 1.0  \\
 German & 130 & 27 & 8 & 0.296 & 45 & 45 & 1.0 \\
 Russian & 105 & 48 & 16 & 0.333 & 0 & 0 & 0  \\
 \textbf{Total} & \textbf{815} & \textbf{750} &  \textbf{261} & \textbf{} & \textbf{437} & \textbf{384} &  \\
 \bottomrule
\end{tabularx}
}
\caption{Number of links and precision values obtained between \lids and other data sets.}
\label{tab:stats}
\end{table*}

Linking \lids to other external resources is based on the string similarities between \lids's resources and the other data sets' resources. The current version of \lids is linked to two other data sets in order to ensure re-usability and integrability. 

The first data set we linked to \lids is \emph{DBnary}. We used the algorithms provided in \limes~\cite{ngomo2012link,sherif2015semantic} framework which are time-efficient to carry out the DBnary linking tasks. The linking was through \texttt{rdfs:label} property using the \texttt{trigram} similarity with acceptance threshold $0.85$.

The second data set we linked with \lids is \emph{BabelNet}. The BabelNet linking process was carried out using the BabelNet API\footnote{\url{http://babelnet.org/guide}} to retrieve senses and definitions. 
While linking, we noticed that \emph{BabelNet} do not correctly type idioms (more details see Section \ref{sec:discussion}). We thus linked to BabelNet manually by comparing our \texttt{skos:definition} property with the \texttt{bn-lemon:de\-fi\-ni\-tion} property of the BabelNet resources. This task was performed by the same group of linguists previously requested.

\subsection{Linking Quality}

In this section, we show and discuss the linking statistics of \lids with BabelNet and DBnary.


\autoref{tab:stats} presents the number of links per resource and language in the \lids data set. Note that all the links were evaluated manually. The \emph{Retrieval} columns show the number of total idioms collected from a given data set and the \emph{Accepted} columns present the number of idioms which were matched exactly as an idiom. We also present the precision achieved by the aforementioned link specifications. 

\emph{DBnary} has presented a good precision in general. Its lower score only comes from Portuguese and Russian as these languages are a bit exploited by linguistic resources in terms of \ac{MWE} thus containing only few idioms. \emph{DBnary} follows the best practice of publishing linked data which means without any typos in labels (e.g \texttt{rdfs:label}) in contrast of \emph{BabelNet} (see \footnote{\url{http://babelnet.org/rdf/page/once_in_a_blue_moon_r_EN}}. This problem contributes for the lower precision score of \emph{BabelNet} because its API does not handle it instead of \limes.


\section{Use Cases}
\label{sec:uses}
In this section, we outline selected application scenarios for our data set. \autoref{lst:lst2}, \autoref{lst:lst3} and \autoref{lst:lst4} illustrate different facets of how \lids can support translation use cases. \lids contains a significant number of instances of concepts, places and translations. Thus, multilingual idioms along with their definitions concerning about a specific information can be easily retrieved from our data set. Moreover, the aligned multilingual representation allows searching for idioms with the same meaning across different languages.

\subsection{Gathering idioms by definitions}
The first use case for our data set is exploratory in nature. 
Machine translation agents are commonly in need of expressions that have a certain meaning. Using a simple SPARQL query over \lids enables these potential agents to easily find idioms which contain a keyword of choice. 
For example, \autoref{lst:lst2} shows a SPARQL query for retrieving English, Italian and Russian idioms which contains the verb \emph{``to deceive''} in their definitions. 

\begin{lstlisting}[label=lst:lst2, float=htb, style=sparql, numbers=left, numberstyle=\tiny, 
caption=Idioms definitions that contains the same verb in (i) English (ii) Italian and (iii) Russian.]
SELECT ?label ?definition
WHERE { 
        ?idiom rdfs:label ?label.
        ?idiom ontolex:sense ?sense.
        ?sense ontolex:isLexicalizedSenseOf ?concept.
        ?concept skos:definition ?definition.
FILTER(bif:contains(?definition, "deceive")) .
FILTER( lang(?label) = "it" || lang(?label) = "en" || lang(?label) = "rus" ).}

\end{lstlisting}

\subsection{Idioms usage per area}
\lids provides information about the place of usage of each idiom. For instance, the idiom \emph{``it's raining cats and dogs"} has English as its language property and comes from England.
By being aware of the place of origin of an idiom, translators are now empowered to translate an idiom to the right idiom for a given target group.  
\autoref{lst:lst3} shows a SPARQL query which retrieves all idioms from England.    

\begin{lstlisting}[label=lst:lst3, float=htb, style=sparql, numbers=left, numberstyle=\tiny, 
caption=All idioms coming from England.]
SELECT ?idiom ?label
WHERE { 
   ?idiom rdfs:label ?label;
             lexvo:usedIn dbr:England .
}
\end{lstlisting}
\subsection{Translating across languages}

Another important use of \lids is to retrieve \emph{indirect translations}. By indirect translation we mean a translation which is based on another translation. Nevertheless, the power of RDF representation of \lids enable the induction of indirect translations through the English translations.
For example, the SPARQL query in \autoref{lst:lst4} first finds the English translation of the German idiom \emph{"Zwei Fliegen mit einer Klappe schlagen"}, then it retrieves Russian idioms with equivalent English translations.

\begin{lstlisting}[label=lst:lst4, float=htb, style=sparql, numbers=left, numberstyle=\tiny, 
caption=Indirect translation.]
SELECT  ?idiom
WHERE { 
  ?i rdfs:label "zwei fliegen mit einer klappe schlagen"@de; 
      ontolex:sense ?sense.  
  ?trans vartrans:source ?sense;
           vartrans:target ?target.
  ?transind vartrans:target ?target;
               vartrans:source ?source.
  ?lex ontolex:sense ?source;
         rdfs:label    ?idiom.
 FILTER( lang(?idiom) = "rus" ). 
}
\end{lstlisting}

\subsection{Third-party uses: Retrieving More Information through Links}

\lids is linked to other data sets, from which we are able to retrieve additional idiom-related information.
For example, \autoref{lst:link} shows a SPARQL query for retrieving a given part-of-speech tag of the English idiom \emph{``out of the blue"} from the same resource exists in \emph{DBnary}.

\begin{lstlisting}[label=lst:link, float=htb, style=sparql, numbers=left, numberstyle=\tiny, 
caption=Retrieving data from different resources.]
SELECT ?pos
    WHERE { 
      ?idiom rdfs:label "out of the blue"@en;
               owl:sameAs ?ext_idiom.
      SERVICE <http://kaiko.getalp.org/sparql> {
        SELECT ?ext_idiom ?pos
        WHERE{
          ?ext_idiom dbnary:partOfSpeech ?pos
        }
    }
}
\end{lstlisting}

\subsubsection{Discussion}
\label{sec:discussion}

A main limitation in the currently available data sets in \ac{LLOD} is the lack of proper categorization of \ac{MWE}. For example, neither \emph{BabelNet} nor \emph{DBnary} have specific \ac{MWE} types. For instance, in \emph{BabelNet}, some idioms were not typed as lexical entries, we were capable of finding exact matches of many idioms which are included in \lids but the matches were from other classes such as a film, a book or music album (e.g., ``head over heels" is the label of a film\footnote{\url{http://babelnet.org/rdf/page/s03412613n}}). In order to alleviate this problem, we also tried to filter the idioms by \emph{bn-lemon:synsetType} in BabelNet, however, incorrect types avoided us to link them easily. For example, the idiom ``The Goose That Laid the Golden Eggs" is typed as ``Named Entity" (see http://babelnet.org/\-rdf/page/s032\-00922n), but it should be a concept. Additionally, \autoref{lst:bn_resources} shows an example resource from BabelNet. 

\begin{lstlisting}[label=lst:bn_resources, float=t, style=sparql, numbers=left, numberstyle=\tiny, 
caption=Fragment of a \emph{BabelNet} resource.]
bn:arm_and_a_leg_n_EN
  a                     lemon:LexicalEntry ;
  rdfs:label            "arm_and_a_leg"@en ;
  lemon:canonicalForm   <http://babelnet.org/rdf/arm_and_a_leg_n_EN/canonicalForm> ;
  lemon:language        "EN" ;
  lemon:sense           <http://babelnet.org/rdf/arm_and_a_leg_EN/s13676929n> ;
  lexinfo:partOfSpeech  lexinfo:noun .
   
\end{lstlisting}

In \autoref{lst:bn_resources}, the idiom \emph{``arm and a leg"} is represented as a \emph{noun} while it should be firstly represented as a \ac{MWE} or more precisely as an idiom. This lack of accurate categorization of \ac{MWE} makes linking data sources such as \lids with other resources very difficult. In particular, using declarative link discovery frameworks for computing similarities among \ac{MWE} without the right classification information becomes a slow task which leads to links with a low level of precision.  

Furthermore, this incomplete categorization exists also in other data sets such as \emph{DBpedia} and \emph{DBnary}. We thus regard \lids as a first effort towards a better LLOD, where \ac{MWE}s (especially idioms) are represented as such. We envision that this better representation will lead to qualitative linked-data driven \ac{NLP} systems, including but not limited to better \ac{MT} applications.

\section{Summary}
\label{sec:conclusion}
In this paper, we described \lids, a multilingual \ac{RDF} data set containing idioms represented in five languages. The data set fills an important gap on \ac{MWE} processing and it can be used as a resource in \ac{NLP} pipelines. The current version of \lids contains $13,889$ triples modeling $815$ concepts with $488$ translations ($115$ indirect translations) coming from $7$ different sources and linked to $645$ external resources. \lids connects idioms from different languages that have semantically equivalent definitions. To ensure interoperability with other data sets on the \ac{LLOD}, \lids is linked to BabelNet and DBnary. 

\subsection{Future Work}

We are currently working to extend the coverage of \lids so that researchers and developers who work on languages not currently present in the data set can benefit from it. Future versions of \lids will include idioms from other languages such as Arabic, Chinese, Korean, Czech, Finnish, and French. Moreover, to handle diatopic language variation, the current languages of \lids are being updated including more fine-grained locations (e.g., cities) as geographical area of use for idioms with more than one meaning even sharing the same country and language. Finally, we plan to improve the automation of the process of internal as well as external linking of idioms by implementing an approach for semantically linking idioms' definitions.

\section*{Acknowledgements} 

We would like to thank all native speakers which contributed with this work. Specifically \emph{Maria Sukha\-reva} from the \emph{Olia} Project\footnote{\url{http://acoli.cs.uni-frankfurt.de/resources/olia/}}, \emph{Will Hanley}\footnote{\url{http://history.fsu.edu/People/Faculty-by-Name/Will-Hanley}}, \emph{Chiara Pace} and \emph{Devayani Bhave}.
Special thanks to \emph{John McCrae}, \emph{Jorge Gracia} and \emph{Gilles S\'{e}rasset} for their constructive discussions about modelling and best practices to create our data set. 

This work has been supported by the H2020 project HOBBIT (GA no. 688227) 
and supported by the Brazilian National Council for Scientific and Technological Development (CNPq) (no. 206971/2014-1). This research has also been supported by the German Federal Ministry of Transport and Digital Infrastructure (BMVI) in the projects LIMBO (no. 19F2029I), OPAL (no. 19F2028A) and GEISER (no. 01MD16014E) as well as by the BMBF project SOLIDE (no. 13N14456).

\section*{Bibliographical References}
\label{main:ref}

\bibliographystyle{lrec}
\bibliography{ref}


\end{document}